\documentclass[10pt,twocolumn,letterpaper]{article}

\usepackage{iccv}
\usepackage{times}
\usepackage{epsfig}
\usepackage{graphicx}
\usepackage{amsmath}
\usepackage{amssymb}
\usepackage{multirow}
\usepackage{makecell}
\usepackage{pifont}
\usepackage{booktabs}
\usepackage{tabularx}

\usepackage[pagebackref=true,breaklinks=true,letterpaper=true,colorlinks,bookmarks=false]{hyperref}

\iccvfinalcopy % *** Uncomment this line for the final submission

\begin{document}

%%%%%%%%% TITLE
\title{CGI-Stereo: Accurate and Real-Time Stereo Matching via Context and Geometry Interaction}

\author{Gangwei Xu\footnotemark[1] \quad\quad Huan Zhou\footnotemark[1] \quad\quad Xin Yang\footnotemark[2]\\
{\normalsize School of EIC, Huazhong University of Science and Technology}\\
{\tt\small \{gwxu, huanzhou, xinyang2014\}@hust.edu.cn}
}

\maketitle
% Remove page # from the first page of camera-ready.
\ificcvfinal\thispagestyle{empty}\fi

%%%%%%%%% ABSTRACT
\begin{abstract}
   In this paper, we propose CGI-Stereo, a novel neural network architecture that can concurrently achieve real-time performance, competitive accuracy, and strong generalization ability. The core of our CGI-Stereo is a Context and Geometry Fusion (CGF) block which adaptively fuses context and geometry information for more effective cost aggregation and meanwhile provides feedback to feature learning to guide more effective contextual feature extraction. The proposed CGF can be easily embedded into many existing stereo matching networks, such as PSMNet, GwcNet and ACVNet. The resulting networks show a significant improvement in accuracy. Specially, the model which incorporates our CGF with ACVNet ranks $1^{st}$ on the KITTI 2012 and 2015 leaderboards among all the published methods. We further propose an informative and concise cost volume, named Attention Feature Volume (AFV), which exploits a correlation volume as attention weights to filter a feature volume. Based on CGF and AFV, the proposed CGI-Stereo outperforms all other published real-time methods on KITTI benchmarks and shows better generalization ability than other real-time methods. Code is available at \textcolor{magenta}{https://github.com/gangweiX/CGI-Stereo}.
   
\end{abstract}

{
\renewcommand{\thefootnote}{\fnsymbol{footnote}}
\footnotetext[1]{Authors contributed equally.}
\footnotetext[2]{Corresponding author.}
}

\section{Introduction}

Stereo matching, which aims to estimate depth (or disparity) from a pair of rectified stereo images, is a fundamental task for many robotics and computational photography applications, such as 3D reconstruction, robot navigation and autonomous driving. Despite a plethora of research works in the literature, state-of-the-art methods still have difficulties in handling repetitive structures, texture-less/transparent objects and occlusions. Meanwhile, concurrently achieving a high accuracy and real-time performance is critical for practical applications yet remains challenging.

\begin{figure}
	\centering	\includegraphics[width=1.0\linewidth]{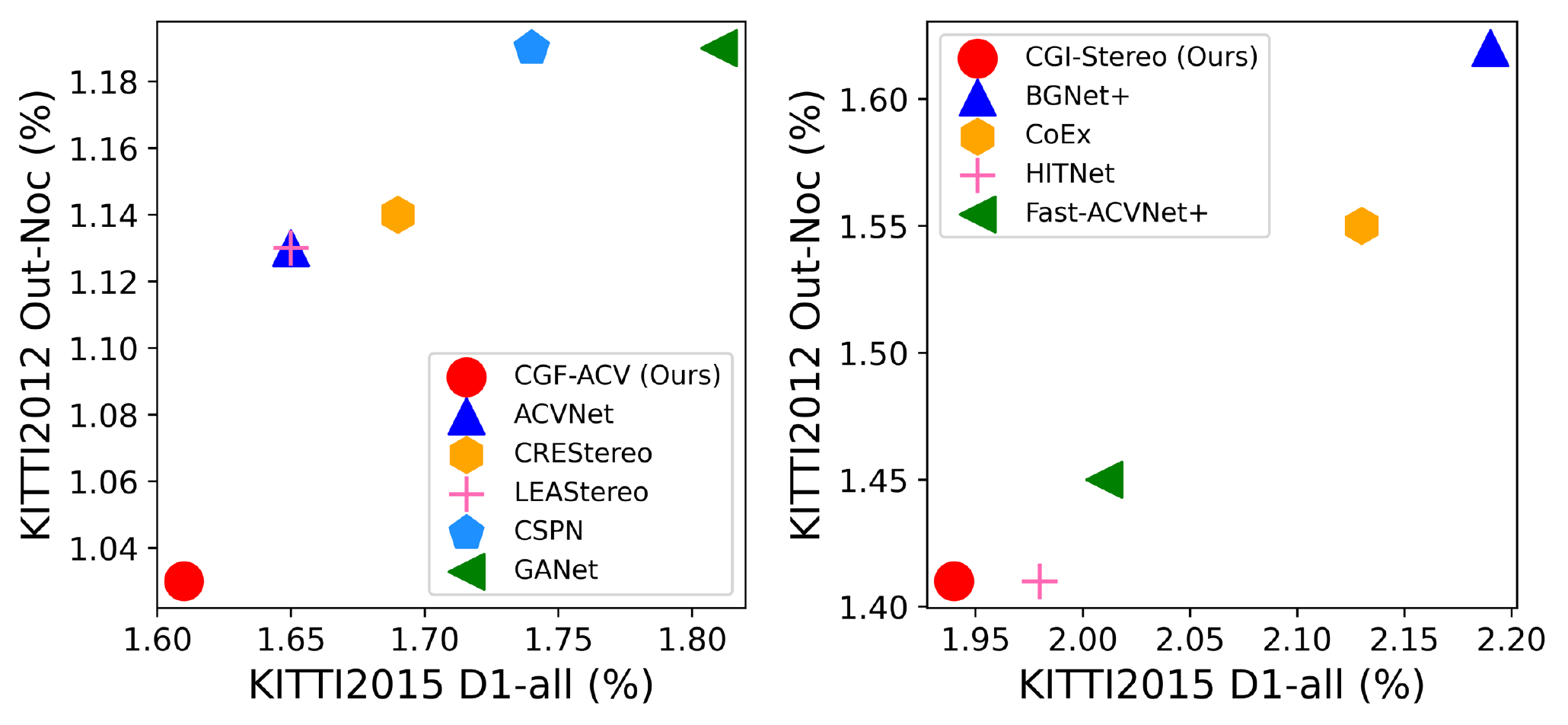}
	\caption{Comparison with state-of-the-art stereo methods. \textbf{Left:} Our CGF-ACV ranks $1^{st}$ on KITTI 2012 and 2015 benchmarks among all the published methods. \textbf{Right:} Our CGI-Stereo outperforms all other published real-time
methods (\textless 50ms) on KITTI benchchmarks.}
	\label{fig:ranking}
 \vspace{-10pt}
\end{figure}

\begin{figure*}
\centering
\includegraphics[width=0.9\textwidth]{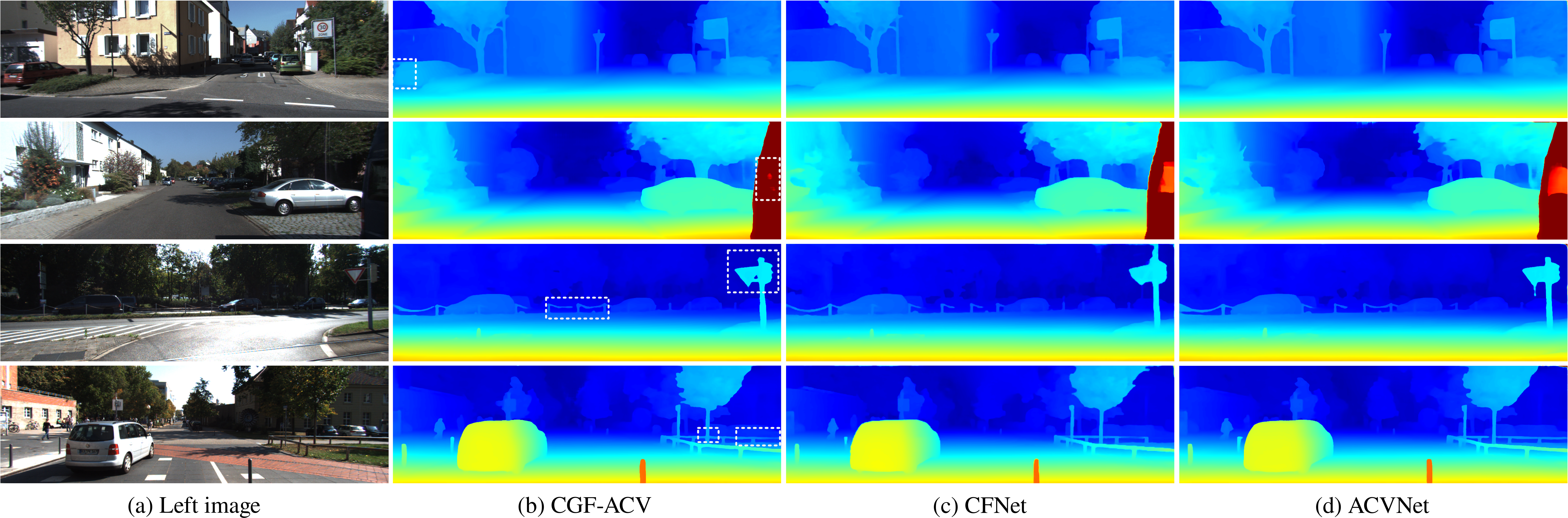} % Reduce the figure size so that it is slightly narrower than the column.
\caption{Visual comparisons with other methods\cite{cfnet,acvnet} on KITTI 2012\cite{kitti2012} and 2015\cite{kitti2015} test set. Our CGF-ACV performs well in occluded regions and reflective surfaces, and preserves more details.}
\label{fig:cgi_acv}
\vspace{-5pt}
\end{figure*}

Recently, stereo matching methods\cite{dispNetC,gcnet,psmnet,gwcnet,acvnet} based on convolutional neural networks (CNNs) have achieved impressive results. State-of-the-art stereo models extract CNN features based on which a cost volume is constructed. The initial cost volume encodes local matching costs yet lacks non-local knowledge, thus it is often ambiguous in occluded regions or large textureless/reflective regions. To address this problem, several cost aggregation networks have been proposed to aggregate contextual matching costs. For instance, GC-Net\cite{gcnet} proposes a 3D encoder-decoder structure to reason about global geometry of the scene from the 4D cost volume. Following GC-Net\cite{gcnet}, PSMNet\cite{psmnet} and GwcNet\cite{gwcnet} design new stacked 3D encoder-decoder structures to regularize the cost volume. These methods demand very deep stacked 3D encoder-decoder layers to aggregate or regularize the cost volume. Although they have shown impressive improvements in accuracy, their computational and memory costs are quite high due to enormous 3D convolutions. To reduce the memory and computational costs, GANet\cite{ganet} designs two guided aggregation layers to replace 3D convolutions. AANet\cite{aanet} proposes intra-scale and inter-scale cost aggregation layers. CoEx\cite{coex} improves cost aggregation by utilizing extracted image features, which excites the cost volume channels with weights computed from the reference image features. Recently developed iterative methods, represented by RAFT-Stereo\cite{raft-stereo}, replace 3D CNN-based aggregation using a recurrent GRU-based updater and repeatedly indexes local costs from the original all-pairs correlation volume to update disparity maps. However, without aggregation the original cost volume is very noisy. As a result, RAFT-Stereo\cite{raft-stereo} demands many GRUs\cite{gru} iterations and a long inference time to obtain a satisfactory disparity map. 

Cost aggregation is critical to accuracy and efficiency in stereo matching yet existing methods require a compromise between accuracy and speed. This paper asks the question, can we design a lightweight 3D encoder-decoder aggregation net to concurrently obtain accurate results and real-time performance? 

Motivated by CoEx\cite{coex} that context features from reference image can guide more accurate understanding of stereo geometry than solely on geometry features, this paper identifies the importance of interaction between contextual features obtained from feature extraction and geometry features encoded in the cost volume and
% rather than purely guiding aggregation using context features, for achieving both a high accuracy and real-time performance.
% The 3D encoder-decoder aggregation structure which is widely-used in most CNN stereo models utilizes stacked down-sampling and up-sampling layers to reduce the computational cost, but this leads to a degradation in accuracy. In fact, it is difficult to decode high resolution geometry features from low resolution geometry features, especially in challenging regions such as occlusion areas, repeated patterns, textureless regions, and reflective surfaces. So, how to fully exploit the potential of 3D encoder-decoder? We observe that context features from reference image features are very beneficial for decoding high-resolution geometry features, rather than relying solely on geometry features. To this end, one motivating question arises: How to interact context and geometry? 
proposes a novel fusion block called Context and Geometry Fusion (CGF).
% which produces spatial attention to adaptively fuse context information at different disparity levels. 
On the one hand, CGF efficiently produces spatial attention weights to adaptively fuse multi-scale context features with geometry features in the cost volume, thus the subsequent 3D decoding layers can decode accurate and high-resolution geometry information with the guidance of context information. On the other hand, the CGF establishes a direct connection between feature extraction and the decoding layers of cost aggregation. By the connection, the context features can be better learned by directly back-propagating the gradients from aggregation and regression to feature extraction. We experimentally show that when truncating the gradient back-propagation flow, i.e. context features only serve as a guidance for geometry feature aggregation, the performance degrades significantly, as shown in Tab.~\ref{tab:interaction_cgf}. We further propose an informative and concise cost volume, named Attention Feature Volume (AFV), which exploits a correlation volume as attention weights to filter a feature volume. 
% Compared to concatenation volume~\cite{gcnet,psmnet}, combined volume\cite{gwcnet} and attention concatenation volume\cite{acvnet}, our AFV can greatly improve the efficiency while maintaining a comparable accuracy (see Tab.~\ref{tab:cv}).

Based on the proposed CGF and AFV, we design an accurate and real-time stereo matching network called CGI-Stereo. At the time of writing, our CGI-Stereo ranks $1^{st}$ on the popular KITTI 2012\cite{kitti2012} and 2015\cite{kitti2015} benchmarks among all the published real-time methods. When trained only on synthetic Scene Flow\cite{dispNetC} dataset, our CGI-Stereo can generalize very well to real datasets such as KITTI 2012 and 2015, ETH3D\cite{eth3d}, and Middlebury\cite{middlebury}, outperforming all other real-time methods. Our CGI-Stereo has also better generalization ability than DSMNet\cite{dsmnet} on KITTI and Middlebury while is $55\times$ faster than it.

In summary, our main contributions are:
\begin{itemize}
    \item We propose Context and Geometry Fusion (CGF) which enables effective interaction between context features and geometry features to greatly improve accuracy. Specially, the model which incorporates our CGF with ACVNet ranks $1^{st}$ on the KITTI 2012 and 2015 benchmarks among all the published methods.
    \vspace{-5pt}
    \item We propose a new cost volume, name Attention Feature Volume (AFV) to encode both matching and content information.
    \vspace{-5pt}
    \item We design an accurate and real-time stereo matching network CGI-Stereo, which outperforms all other published real-time methods on KITTI benchmarks and shows better generalization ability than other real-time methods.
    \vspace{-5pt}
\end{itemize}

\begin{figure*}[t]
    \centering
    \includegraphics[width=0.9\linewidth]{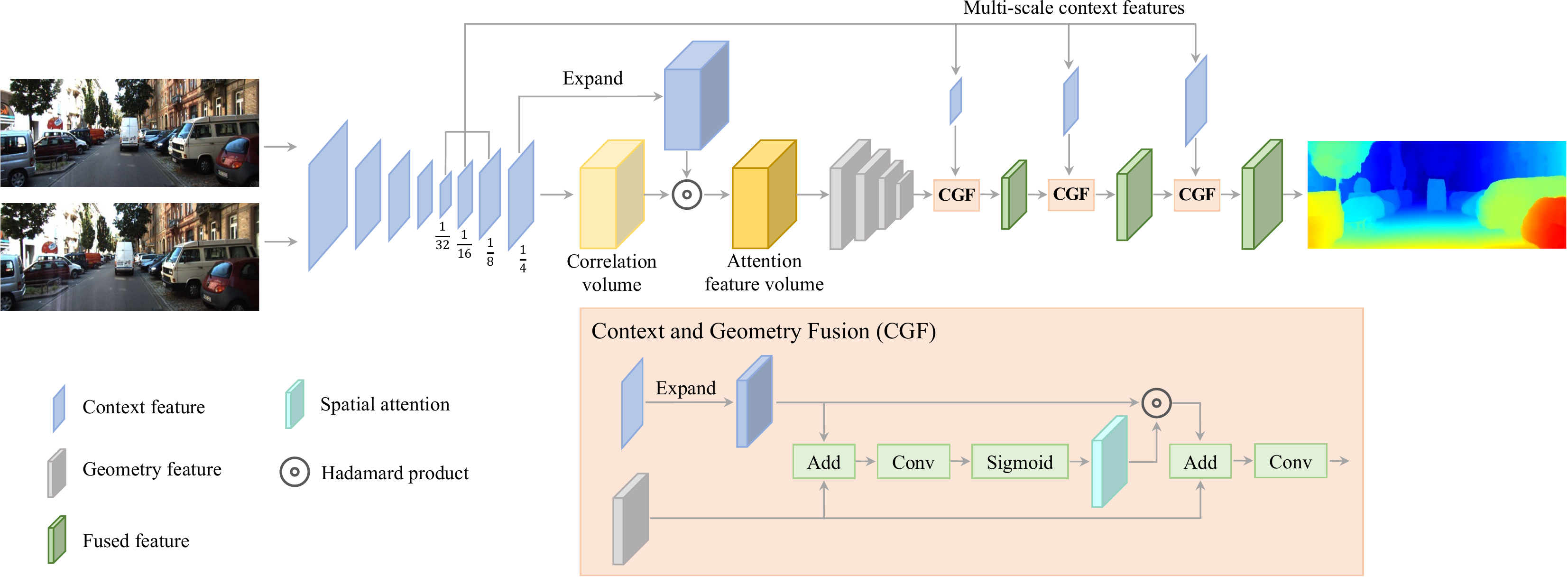}
    \vspace{10pt}
    \caption{Overview of our proposed CGI-Stereo. To improve the effectiveness and efficiency of aggregation, we propose Context and Geometry Fusion (CGF) which produces spatial attention weights to adaptively fuse context information and geometry information. We also construct an informative and concise cost volume, named Attention Feature Volume (AFV), which exploits a correlation volume as attention weights to filter a feature volume.}
    \label{fig:cgistereo}
    \vspace{-10pt}
\end{figure*}

\section{Related Work}
\label{sec:related_work}
\textbf{Stereo Matching}. Recently, CNNs have shown great potential for stereo matching tasks. To handle complex real world scenes, especially texture-less regions and occlusions, popular stereo methods\cite{psmnet,gwcnet,nie2019multi,cfnet,acvnet,acfnet,multilevel,hierarchical} usually use 2D convolutions to extract robust features based on which a cost volume is constructed. 
% Then they use a large number of 3D convolutions to regularize the cost volume. 
GC-Net\cite{gcnet} constructs a 4D cost volume by concatenating the left and right CNN features and then utilizes a 3D encoder-decoder network to aggregate and regularize the cost volume. Following GC-Net\cite{gcnet}, PSMNet\cite{psmnet} and GwcNet\cite{acvnet} exploit the stacked 3D encoder-decoder structure to regularize cost volume, achieving great improvement in accuracy. However, the massively stacked 3D convolution layers are computationally expensive and memory-consuming. To improve efficiency of cost aggregation, GANet\cite{ganet} proposes two guided aggregation layers to replace abundant 3D convolutions, and AANet\cite{aanet} proposes intra-scale and inter-scale aggregation layers. Without 3D convolutions, RAFT-Stereo\cite{raft-stereo} exploits GRUs to iteratively update disparity maps from all-pairs correlation costs. However, these methods still have difficulties achieving high accuracy and real-time performance concurrently.

Several studies\cite{deeppruner,bgnet,fadnet, decomposition,hitnet,fast-acv} focus on lightweight stereo networks to achieve real-time performance and meanwhile maintain satisfactory accuracy. They usually construct a low-resolution cost volume or a sparse cost volume to greatly reduce the amount of computation. For example, StereoNet\cite{stereonet} and BGNet\cite{bgnet} construct and aggregate a low-resolution cost volume, and then use an edge-preserving refinement module and an up-sampling module based on the learned bilateral grid respectively to improve accuracy. DeepPruner\cite{deeppruner} prunes the disparity search range to build a sparse cost volume. DecNet\cite{decomposition} runs dense matching at a very low resolution and uses sparse matching at high resolution. Fast-ACV\cite{fast-acv} proposes to exploit a low-resolution correlation volume to generate disparity hypotheses with high likelihood and the corresponding attention weights, and then construct sparse attention concatenation volume. Although these methods achieve real-time performance, they lead to great degradation in accuracy. Different from existing methods, our proposed CGF can adaptively fuse context features into geometry features, which helps to decode accurate high-resolution geometry features. The proposed CGI-Stereo achieves both real-time performance and high accuracy, even surpassing some top-performing methods\cite{psmnet,gwcnet}.

\textbf{Context and Geometry Fusion for Stereo Matching}. Context is crucial in reasoning about the geometry of a scene. To combine context and geometry, GC-Net\cite{gcnet} constructs a 4D concatenation volume by concatenating left and right context features across all disparity levels and relies on the subsequent 3D convolutions for information fusion and mining. SegStereo\cite{segstereo} embeds semantic features from segmentation into the cost volume to improve accuracy. EdgeStereo\cite{edgestereo} integrates edge cues into the cost volume to preserve subtle details. In contrast, our CGF adaptively fuses context and geometry information for more effective cost aggregation and  meanwhile provides feedback to feature learning to guide more effective contextual feature extraction. Our method outperforms GC-Net, SegStereo and EdgeStereo by a large margin in terms of accuracy and speed.

\section{Method}

As illustrated in Fig.~\ref{fig:cgistereo}, our CGI-Stereo consists of multi-scale feature extraction, attention feature volume construction, cost aggregation and disparity prediction. The cost aggregation includes a 3D encoder and a decoder based on context and geometry fusion (CGF). We exploit CGF to decode accurate and high-resolution geometry features from a low-resolution cost volume. In this section, we first describe the design of our CGF (Sec.~\ref{sec:cgf}). Then in Sec.~\ref{sec:network}, we present details of the network architecture of CGI-Stereo. Finally, in Sec.~\ref{sec:loss}, we describe the loss functions used to train our CGI-Stereo.

\begin{figure}
	\centering	\includegraphics[width=1.0\linewidth]{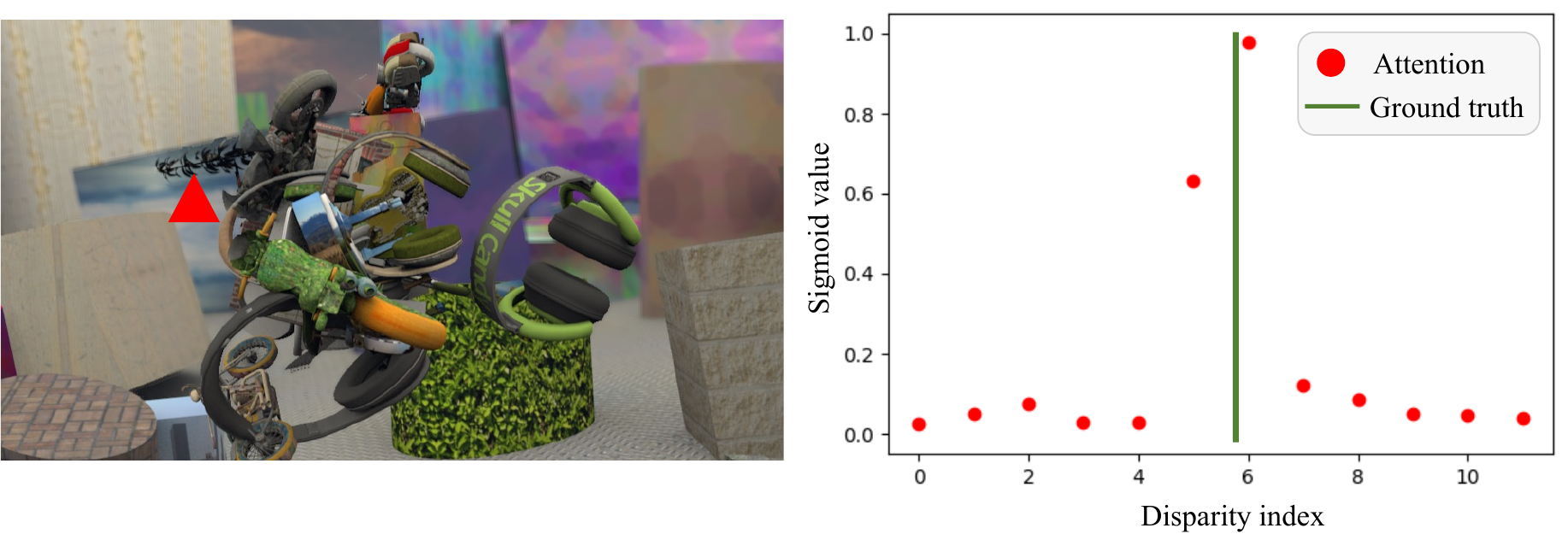}
	\caption{We display the attention in CGF for the red triangle position of the input image along the disparity dimension at 1/16 resolution. The CGF can adaptively select `important' regions for context and geometry fusion by attention.}
	\label{fig:att_cgf}
 \vspace{-10pt}
\end{figure}

\subsection{Context and Geometry Fusion} \label{sec:cgf}
To decode accurate and high-resolution geometry information from low-resolution geometry with assistance of context information, we propose Context and Geometry Fusion (CGF) for effective and flexible cost aggregation.

Given context features $\mathbf{C}\in\mathbb{R}^{B\times{C_0}\times{H_0}\times{W_0}}$ from left image (obtained by feature extraction module denoted by light blue squares in Fig.~\ref{fig:cgistereo}) and geometry features $\mathbf{G}\in\mathbb{R}^{B\times{C_0}\times{D_0}\times{H_0}\times{W_0}}$ (obtained by applying 3D convolutions to AFV denoted using gray cubes in Fig.~\ref{fig:cgistereo}), we expand the size of $\mathbf{C}$ to ${B\times{C_0}\times{D_0}\times{H_0}\times{W_0}}$ ($B$: batch, $C_0$: channel, $D_0$: disparity, $H_0$: height, $W_0$: width), denoted as $\mathbf{C}_{expand}$. Rather than adding $\mathbf{C}_{expand}$ and $\mathbf{G}$, we use an attention mechanism to fuse features. Geometry features tend to have a unimodal distribution in the disparity dimension. Thus directly adding expanded context features to geometry features, that is, context features are shared across the disparity channels, completely ignores the differences in disparity dimension and in turn yields an ineffective fusion. Inspired by \cite{cbam}, we generate spatial attention weights by exploring the spatial relationship of context and geometry features. The spatial attention weights can adaptively select `important' regions for context and geometry fusion (see Fig. \ref{fig:att_cgf}). To compute the spatial attention weights, we first sum $\mathbf{C}_{expand}$ and $\mathbf{G}$ and then apply a convolution operation as,
\begin{equation}
\mathbf{A}_{s}=\sigma({f}^{5\times5}(\mathbf{G}+\mathbf{C}_{expand})),
\label{equ:spa_att}
\end{equation}
where $\sigma$ represents the sigmoid function and ${f}^{5\times5}$ denotes a convolution operation with the filter size of $1\times5\times5$. The spatial attention weights $\mathbf{A}_{s}\in\mathbb{R}^{B\times{C_0}\times{D_0}\times{H_0}\times{W_0}}$ encodes where to emphasize or suppress. Accordingly, we fuse $\mathbf{C}_{expand}$ and $\mathbf{G}$ as,
\begin{equation}
\mathbf{G}_{fused}={f}^{5\times5}(\mathbf{G}+\mathbf{A}_{s}\odot \mathbf{C}_{expand}),
\label{equ:fused}
\end{equation}
where $\odot$ denotes the Hadamard Product.

\subsection{Network Architecture of CGI-Stereo} \label{sec:network}
\subsubsection{Multi-scale Feature Extraction}
Given an input stereo image pair whose size is $H\times{W}\times3$, we use a pretrained MobileNetV2 on ImageNet\cite{imagenet} as our backbone to obtain four scales of feature maps whose resolutions are 1/4, 1/8, 1/16, and 1/32 of the original resolution respectively. Following CoEx\cite{coex},  we repeatedly perform up-sampling on the feature map, until its size reaches $H/4 \times W/4$. In more details, each up-sampling block applies a transpose convolution with kernel $4 \times 4$ and stride 2 to up-sample feature map of coarser resolution. Features are concatenated with skip-connection, and then a $3 \times 3$ convolution is applied to merge the skipped and upsampled features for the current resolution. Finally, we obtain multi-scale context features which are then used for correlation volume construction, AFV construction and CGF as shown in Fig.~\ref{fig:cgistereo}.

\subsubsection{Attention Feature Volume Construction}
We construct an informative and concise cost volume, named Attention Feature Volume (AFV), which exploits correlation volume as attention weights to filter a feature volume. Firstly, we exploit the feature maps at 1/4 resolution to construct a correlation volume $\mathbf{V}_{corr}$ as, 
\begin{equation}
\mathbf{V}_{corr}(:,d,x,y)=\frac{<\mathbf{f}_{l}(:,x,y), \mathbf{f}_{r}(:,x-d,y)>}{{||\mathbf{f}_{l}(:,x,y)||}_2\cdot{||\mathbf{f}_{r}(:,x-d,y)||}_2},
\end{equation}
where $d$ is disparity index, and $(x, y)$ represents the pixel coordinate. $\mathbf{f}_{l}$ and $\mathbf{f}_{r}$ are the left and right feature maps. The calculated correlation volume by cosine similarity has only one channel, thus we perform a $3 \times 3$ convolution followed by a BatchNorm and a leaky ReLU to increase its channel to 8, denoted as $\mathbf{A}_{corr}\in\mathbb{R}^{B \times C \times D/4 \times H/4 \times W/4}$ $(C=8)$. Then we expand the size of the left feature maps at 1/4 resolution to $B \times C \times D/4 \times H/4 \times W/4$, denoted as $\mathbf{F}_{l}$. Finally, the attention feature volume $\mathbf{V}_{AF}$ is computed as,
\begin{equation}
\mathbf{V}_{AF}=\mathbf{A}_{corr}\odot \mathbf{F}_{l}.
\label{equ:att_vol}
\end{equation}

The attention feature volume encodes both matching and context information. Compared with combined volume proposed by~\cite{gwcnet} and attention concatenation volume proposed by~\cite{acvnet}, our attention feature volume is more efficient. As both combined volume and attention concatenation volume require additional cost to construct the concatenation volume.

\subsubsection{Cost Aggregation}
To decode accurate and high-resolution geometry features, we propose CGF to fuse multi-scale context features with geometry features (see Fig.~\ref{fig:cgistereo}). First, we exploit three down-sampling modules to aggregate matching features and increase features' receptive field while reducing computation. Each down-sampling module consists of a $3\times3\times3$ 3D convolution with stride 2 and a $3\times3\times3$ 3D convolution with stride 1. After down-sampling, the size of geometry features is $B \times 6C \times D/32 \times H/32 \times W/32$. Then we alternately use CGF and an up-sampling module to decode high-resolution geometry features. Each up-sampling module consists of a $4\times4\times4$ 3D transposed convolution with stride 2 and two $3\times3\times3$ 3D convolutions with stride 1.

\subsubsection{Disparity Prediction}
For the aggregated cost volume, we pick out the top 2 values at every pixel following CoEx\cite{coex}, and perform softmax on these values to compute the expected disparity. The computed disparity map $\mathbf{d}_0$ has the size of $B \times 1 \times H/4 \times W/4$. Then we exploit `superpixel' weights surrounding each pixel as \cite{yang2020superpixel} to up-sample disparity map $\mathbf{d}_0$ to the original resolution $\mathbf{d}_1 \in \mathbb{R}^{B \times 1 \times H \times W}$.

\subsection{Loss Function} \label{sec:loss}
The whole network is trained in a supervised end-to-end manner. The final loss function is given by,
\begin{equation}
    \mathcal{L} = \lambda_{0}Smooth_{L_1}(\mathbf{d}_0-\mathbf{d}_{gt})+\lambda_{1}Smooth_{L_1}(\mathbf{d}_1-\mathbf{d}_{gt})
\end{equation}
where $\mathbf{d}_{0}$ is disparity map of 1/4 resolution, $\mathbf{d}_{1}$ is the final disparity map at full resolution. The $\mathbf{d}_{gt}$ denotes the ground-truth disparity map.

\begin{table*}
    \centering
    \begin{tabular}{l|cc|cccccc}
     \toprule
     {Method} & AFV & CGF & EPE (px) & D1 (\%) & \textgreater 1px (\%) & \textgreater 2px (\%) & \textgreater 3px (\%) & Time (ms) \\
     \midrule
     Baseline &\ding{55} & \ding{55} &0.82 &3.08 &9.67 &5.20 &3.76 &27 \\
     AFV & \ding{51} &\ding{55} &0.74 &2.67 &8.31 &4.51 &3.28 &28 \\
     CGF &\ding{55} &\ding{51} &0.66 &2.34 &7.47 &4.03 &2.92 &28 \\
     AFV+CGF (CGI-Stereo) &\ding{51} &\ding{51} &\textbf{0.64} &\textbf{2.24} &\textbf{7.17} &\textbf{3.86} &\textbf{2.80} &29 \\
     
    \bottomrule
    \end{tabular}
    \vspace{10pt}
    \caption{Ablation study on Scene Flow\cite{dispNetC}.}
\label{tab:ablation}
% \vspace{-5pt}
\end{table*}

\section{Experiment}
\subsection{Datasets and Evaluation Metrics}
\textbf{Scene Flow}\cite{dispNetC} is a collection of synthetic stereo datasets which provides 35,454 training image pairs and 4,370 testing image pairs with the resolution of 960×540. This dataset provides dense disparity maps as ground truth. For Scene Flow, we utilize the widely-used evaluation metrics the end point error (EPE) and the percentage of disparity outliers D1 as the evaluation metrics. The outliers are defined as the pixels whose disparity errors are greater than max($3px,\; 0.05\mathbf{d}_{gt}$), where $\mathbf{d}_{gt}$ denotes the ground-truth disparity.

\textbf{KITTI} includes KITTI 2012\cite{kitti2012} and KITTI 2015\cite{kitti2015}, which are datasets for real-world driving scenes. KITTI 2012 contains 194 training pairs and 195 testing pairs, and KITTI 2015 contains 200 training pairs and 200 testing pairs. Both datasets provide sparse ground-truth disparities obtained with LIDAR. For KITTI 2012, we report the percentage of pixels with errors larger than x disparities in both non-occluded (x-noc) and all regions (x-all), as well as the overall EPE in both non occluded (EPE-noc) and all the pixels (EPE-all). For KITTI 2015, we report the percentage of pixels with EPE larger than 3 pixels in background regions (D1-bg), foreground regions (D1-fg), and all (D1-all). We also use the training sets of KITTI 2012 and KITTI 2015 for generalization performance evaluation. The \textgreater3px (i.e., the percentage of points with absolute error larger than 3 pixel) is reported.

\textbf{Middlebury 2014}\cite{middlebury} is an indoor dataset with 15 training image pairs and 15 testing image pairs with full, half, and quarter resolutions. We use the training image pairs with half resolution to evaluate the cross-domain generalization performance. \textgreater2px (i.e., the percentage of points with absolute error larger than 2 pixel) is reported.

\textbf{ETH3D}\cite{eth3d} is a collection of grayscale stereo pairs from indoor and outdoor scenes. It contains 27 training and 20 testing image pairs with sparse labeled ground-truth. The training set is used to evaluate cross-domain generalization performance. We report \textgreater1px (i.e., the percentage of points with absolute error larger than 1 pixel) metric.

% \begin{table}
%     \centering
%     \begin{tabular}{l|cccc}
%      \toprule
%      Method
%      & \makecell{EPE \\ (px)} & \makecell{D1 \\ (\%)} & \makecell{\textgreater 3px \\ (\%)} & \makecell{Time \\ (ms)} \\
%      \midrule
%      Combined volume\cite{gwcnet} &0.65 &2.29 &2.86 &36 \\
%      ACV\cite{acvnet} &0.64 &2.24 &2.81 &35 \\
%      AFV (Ours) &0.64 &2.24 &2.80 &29 \\     
%     \bottomrule
%     \end{tabular}
%     \vspace{10pt}
%     \caption{Cost volume analysis. Compared to combined volume\cite{gwcnet} and ACV\cite{acvnet}, our AFV is more efficient while maintaining a comparable accuracy.}
% \label{tab:cv}
% \end{table}

\subsection{Implementation Details}
We implement our methods with PyTorch and perform our experiments using NVIDIA RTX 3090 GPUs. For all the experiments, we use the Adam\cite{adam} optimizer, with $\beta_1=0.9$, $\beta_2=0.999$. The coefficients of two outputs are set as $\lambda_{0}$=0.3, $\lambda_{1}$=1.0. On Scene Flow\cite{dispNetC}, we first train CGI-Stereo network for 20 epochs, and then fine-tune it for another 20 epochs. The initial learning rate is set to 0.001 decayed by a factor of 2 after epoch 10, 14, 16, and 18. For KITTI, we fine-tune the pre-trained model on Scene Flow for 600 epochs on the mixed KITTI 2012\cite{kitti2012} and KITTI 2015\cite{kitti2015} training sets. The initial learning rate is 0.001 and decreases to 0.0001 at the $300^{th}$ epoch.

\subsection{Ablation Study}
To validate the effectiveness of AFV and CGF proposed in this paper, we conduct ablation experiments on the Scene Flow test set. We integrate AFV and CGF into a baseline model. The baseline model constructs a correlation volume and uses a common 3D encoder-decoder structure to regularize the correlation volume. As shown in Tab.~\ref{tab:ablation}, our proposed AFV and CGF can significantly improve performance, especially CGF, which improves EPE metric from 0.82 to 0.66. The best performance is obtained by integrating AFV and CGF simultaneously, which is denoted as CGI-Stereo.

\subsection{Analysis}
% \subsubsection{Attention Feature Volume}
% We compare our attention feature volume (AFV) with attention concatenation volume (ACV) proposed by ACVNet\cite{acvnet} and combined volume proposed by GwcNet\cite{gwcnet}. For all comparison models in this study, only the cost volume construction is different, other components remain the same with CGI-Stereo. Compared with the combined volume and ACV which construct a concatenation volume to encode context information, our feature volume of AFV already encodes sufficient context information. Results in Tab.~\ref{tab:cv} show that, our AFV is more efficient while maintaining comparable accuracy compared with the top-performing cost volumes.

% \begin{table}
% \begin{center}
% \begin{tabular}{c|c|c}
% \toprule
% Method &Memory (GB) &Time (ms) \\ 
% \midrule
% w/o CGF & 0.33 & 28 \\
% full method & 0.33 & 29 \\
% \bottomrule
% \end{tabular}
% \end{center}
% \caption{Memory cost and inference time analysis on KITTI image resolution.}\label{tab:memory_cost}
% \vspace{-10pt}
% \end{table}

\begin{table}
    \centering
    \begin{tabular}{cc|cccc}
     \toprule
     \multicolumn{2}{c|}{Position of CGF} & EPE & D1 & \textgreater3px & Time\\
     Encoder & Decoder & (px) & (\%) & (\%) & (ms) \\
     \midrule
     \ding{55} &\ding{55} &0.74 &2.67 & 3.28 &28 \\
     \ding{51} &\ding{55}  &0.70 &2.48 &3.09 &29 \\
     \ding{55}& \ding{51} &0.64 &2.24 &2.80 &29 \\
     \ding{51} & \ding{51} &0.64 &2.25 &2.82 &30 \\
     
    \bottomrule
    \end{tabular}
    \vspace{10pt}
    \caption{Position of CGF. Significant
improvement is obtained when integrating CGF into the 3D decoder structure.}
\label{tab:posi_cgf}
% \vspace{-5pt}
\end{table}

\textbf{Position of CGF.} We evaluate the performance of CGF at different positions in the 3D encoder-decoder structure, as shown in Tab.~\ref{tab:posi_cgf}. When using CGF in the 3D encoder structure, the improvement is very slight. While significant improvement is obtained when integrating it into a 3D decoder structure. Our analysis shows that our AFV contains sufficient context information which can be used by the 3D encoder to extract context information. However, the 3D decoder is difficult to accurately recover high-resolution geometry information from low-resolution geometry information. Thus CGF can exploit context knowledge as effective guidance. In addition, through CGF operations, context features can be directly supervised, which helps to learn better feature representations.

\begin{table}
    \centering
    \begin{tabular}{l|cccc}
     \toprule
     Method
     & \makecell{EPE \\ (px)} & \makecell{D1 \\ (\%)} & \makecell{\textgreater 3px \\ (\%)} & \makecell{Time \\ (ms)} \\
     \midrule
     Truncating gradient &0.72 &2.46 &3.10 &29 \\
     No Truncating &\textbf{0.64} &\textbf{2.24} &\textbf{2.80} &29 \\
    \bottomrule
    \end{tabular}
    \vspace{10pt}
    \caption{Interaction analysis of CGF.}
\label{tab:interaction_cgf}
\vspace{-10pt}
\end{table}

\textbf{Interaction analysis of CGF.} The CGF which interacts context and geometry can facilitate the learning of contextual and geometric features. As shown in Tab.~\ref{tab:interaction_cgf}, when truncating the gradient back-propagation flow of contextual features in CGF, that is, the contextual features only serve as a guide for geometric features, the performance degrades significantly. While by CGF with the gradient back-propagation flow (without truncating), geometrical features can also affect the learning process of contextual features and improve  the effectiveness contextual features.

\subsection{Performance of CGF}
To demonstrate the superiority of our CGF, we integrate our CGF into three state-of-the-art models, i.e. PSMNet\cite{psmnet}, GwcNet\cite{gwcnet}, and ACVNet\cite{acvnet}. We compare the performance of the original models with those after using our CGF. As shown in Tab.~\ref{tab:performance_cgi},  our CGF can improve the performance of the original methods on KITTI benchmarks by a large margin. Specially, our CGF improves PSMNet by 18.8\%, GwcNet by 11.4\%, and ACVNet by 8.8\% on KITTI 2012\cite{kitti2012} for the 3-noc metric. And for D1-all metric on KITTI 2015\cite{kitti2015}, our CGF can improve PSMNet by 22.4\%, GwcNet by 19.0\%.

\begin{table} 
\footnotesize
\setlength{\tabcolsep}{4.pt} %
    \centering
    \begin{tabular}{l|cccc|ccc}
     \toprule
     \multirow{2}{*}{Method} & \multicolumn{4}{c|}{KITTI 2012\cite{kitti2012}} &\multicolumn{3}{c}{KITTI 2015\cite{kitti2015}}\\
     & 3-noc & 3-all & 4-noc & 4-all &D1-bg & D1-fg & D1-all \\
     \midrule
     PSMNet\cite{psmnet} &1.49 &1.89 &1.12 &1.42 &1.86 & 4.62 &2.32 \\
     CGF-PSM &\textbf{1.21} &\textbf{1.57} &\textbf{0.91} &\textbf{1.18} &\textbf{1.46} &\textbf{3.47} &\textbf{1.80} \\
     \midrule
     GwcNet\cite{gwcnet} &1.32 &1.70 &0.99 &1.27 &1.74 &3.93 &2.11 \\
     CGF-Gwc &\textbf{1.17} &\textbf{1.52} &\textbf{0.89} &\textbf{1.15} &\textbf{1.38} &\textbf{3.34} &\textbf{1.71} \\
     \midrule
     ACVNet\cite{acvnet} &1.13 &1.47 &0.86 &1.12 &1.37 & \textbf{3.07} &1.65 \\
     CGF-ACV &\textbf{1.03} &\textbf{1.34} &\textbf{0.78} &\textbf{1.01} &\textbf{1.31} &3.08 &\textbf{1.61} \\
     
    \bottomrule
    \end{tabular}
    \vspace{10pt}
    \caption{Performance of CGF. Our CGF improves the performance of the state-of-the-art methods\cite{psmnet,gwcnet,acvnet} on KITTI benchmarks by a large margin. \textbf{Bold}: Best.}
\label{tab:performance_cgi}
% \vspace{-5pt}
\end{table}

\subsection{Comparisons with State-of-the-art}
\textbf{Scene Flow.} As shown in Tab.~\ref{tab:scene_flow}, our CGI-Stereo outperforms all other real-time methods\cite{bgnet, decomposition, coex} (i.e., the inference time for a stereo pair is less than 50ms), which achieves the remarkable EPE of 0.64. In addition, our CGI-Stereo also outperforms some complex stereo models, including PSMNet\cite{psmnet}, GwcNet\cite{gwcnet}, GANet\cite{ganet}, LEAStereo\cite{leastereo} and CFNet\cite{cfnet}.

\begin{table}
    \centering
    \begin{tabular}{l|cc}
     \toprule
     Method
     & EPE (px) & Runtime (ms) \\
     \midrule
     PSMNet\cite{psmnet} &1.09 &410 \\
     GwcNet\cite{gwcnet} &0.76 &320 \\
     GANet\cite{ganet} &0.84 &360 \\
     LEAStereo\cite{leastereo} &0.78 &300 \\
     CFNet\cite{cfnet} &0.97 &180 \\
     StereoNet\cite{stereonet} &1.10 &15 \\
     BGNet\cite{bgnet} &1.17 &25 \\
     DecNet\cite{decomposition} &0.84 &50 \\
     CoEx\cite{coex} &0.68 &27 \\
     CGI-Stereo (Ours) &\textbf{0.64} &29 \\
     
    \bottomrule
    \end{tabular}
    \vspace{10pt}
    \caption{Comparison with the state-of-the-art on Scene Flow\cite{dispNetC}.}
\label{tab:scene_flow}
\vspace{-15pt}
\end{table}

\begin{figure}
	\centering	\includegraphics[width=0.9\linewidth]{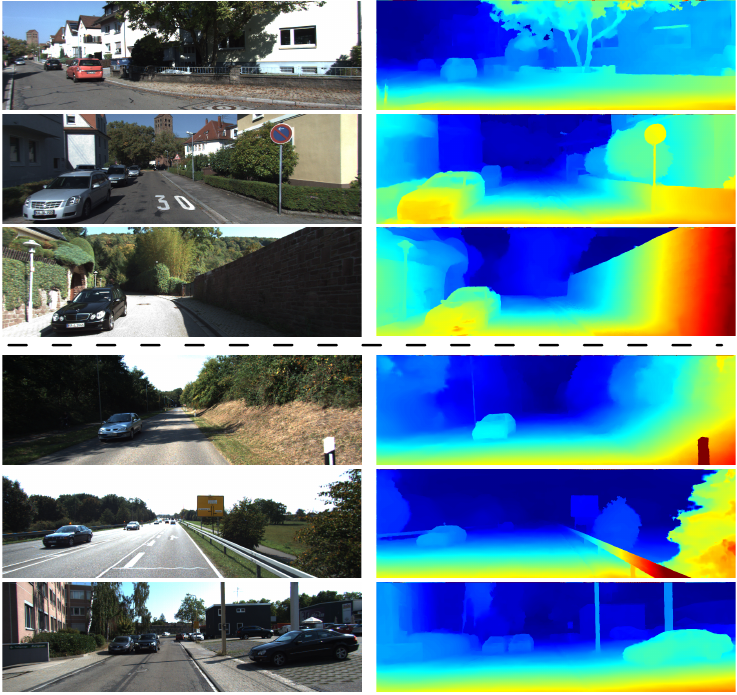}
	\caption{Generalization results of CGI-Stereo on KITTI 2012 and 2015.}
	\label{fig:genera_kitti}
 \vspace{-10pt}
\end{figure}

\begin{figure*}
\centering
\includegraphics[width=0.8\textwidth]{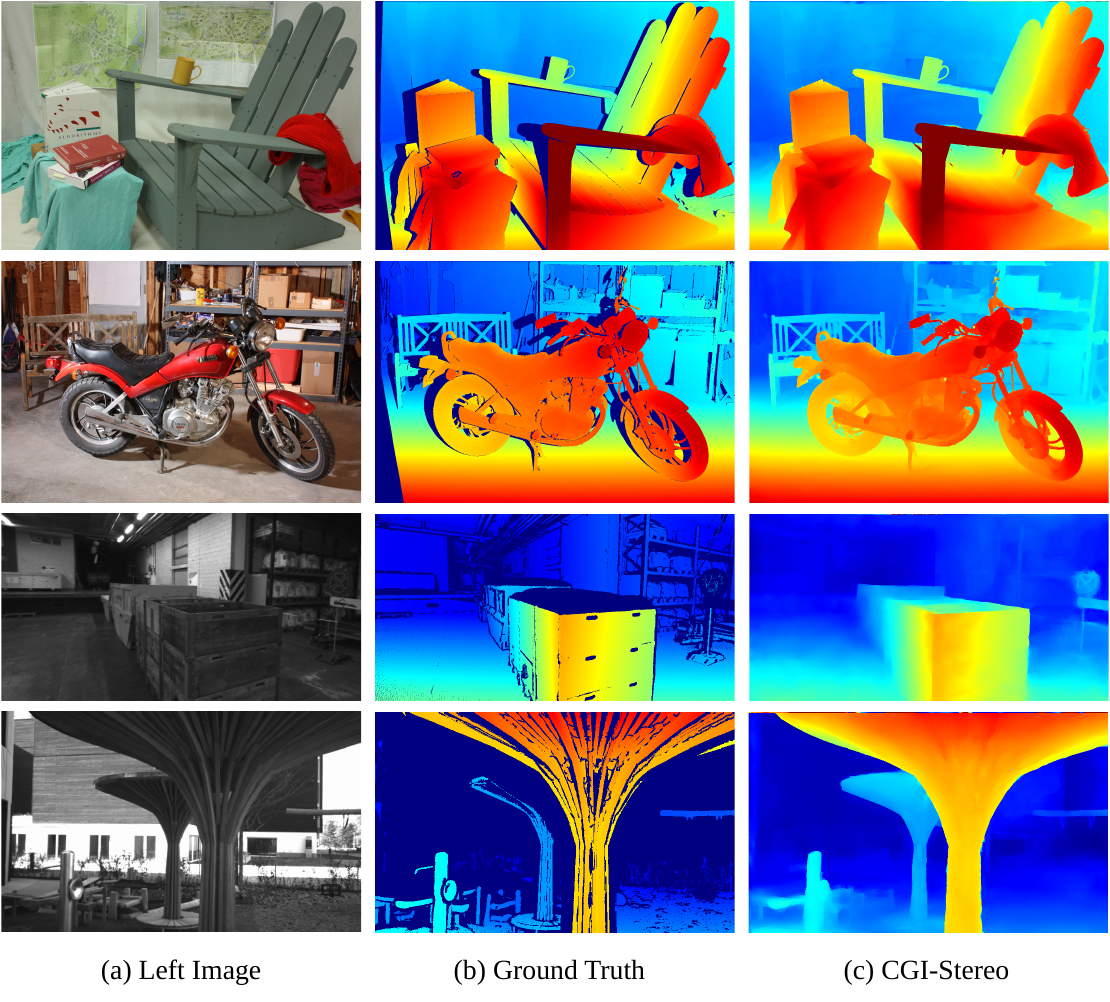} % Reduce the figure size so that it is slightly narrower than the column.
\caption{Generalization results on Middlebury 2014\cite{middlebury} and ETH3D\cite{eth3d}. Our method generalizes well to real-world scenarios when only trained on synthetic Scene Flow dataset\cite{dispNetC}.}
\label{fig:genera_midd_eth}
\vspace{-10pt}
\end{figure*}

\begin{table*}
    \centering
    \begin{tabular}{c|l|cccccc|ccc|c}
    \toprule
     \multirow{2}{*}{Target} & \multirow{2}{*}{Method} & \multicolumn{6}{c|}{KITTI 2012\cite{kitti2012} } & \multicolumn{3}{c|}{ KITTI 2015\cite{kitti2015}}  &\\
    & & 3-noc & 3-all & 4-noc & 4-all & \thead{EPE \\ noc} & \thead{EPE\\all} & D1-bg & D1-fg & D1-all &\makecell{Runtime \\ (ms)}  \\
    \midrule
    
    \multirow{11}{*}{ \rotatebox{90}{\textit{Accuracy}}}
    % & {PSMNet\cite{psmnet}}  &1.49  & 1.89 & 1.12 & 1.42  & 0.5 & 0.6 &1.86  &4.62  &2.32  & 410\\
    % & {GwcNet\cite{gwcnet}}  & 1.32 & 1.70 & 0.99 & 1.27 & 0.5  & 0.5 & 1.74 & 3.93 & 2.11  & 320\\ 
    & {GANet\cite{ganet}}  & 1.19 & 1.60 & 0.91 & 1.23 & 0.4  & 0.5 & 1.48 & 3.46 & 1.81  & 1800\\ 
    % & DeepPrunerBest~\cite{duggal2019deeppruner} & - & - & - & - & - & -  &1.87 &3.56 &2.15 &182 \\
    & LaC+GANet\cite{lsp} & 1.05 &1.42 &0.80 &1.09 &0.4 &0.5 &1.44 & \textbf{2.83} & {1.67} &1800 \\ 
    & CFNet\cite{cfnet} & 1.23 & 1.58 & 0.92 & 1.18 & 0.4 & 0.5  &1.54 &3.56 &1.88 &180 \\
    & SegStereo\cite{segstereo} & 1.68 &2.03 &1.25 &1.52 &0.5 &0.6 &1.88 &4.07 &2.25 &600\\
    &SSPCVNet \cite{sspcv} & 1.47 &1.90 &1.08 &1.41 &0.5 &0.6 &1.75 &3.89 &2.11 &900 \\
    & EdgeStereo-V2\cite{edgestereo} &1.46 &1.83 &1.07 &1.34 &0.4 &0.5 & 1.84 &3.30 &2.08 &320 \\
    & CSPN\cite{cspn} &1.19 &1.53 &0.93 &1.19 &- &- &1.51 & 2.88 &1.74 & 1000 \\
    & LEAStereo\cite{leastereo} & {1.13} & {1.45} & {0.83} & {1.08} & 0.5 & 0.5  &1.40 &{2.91} &1.65 &300\\
    & CREStereo\cite{crestereo} &1.14 &1.46 &0.90 &1.14 &0.4 & 0.5 & 1.45 &2.86 &1.69 &410\\
    & ACVNet & 1.13 & 1.47 & 0.86 & 1.12 & 0.4 & 0.5  &1.37 &3.07 &1.65 &200 \\
    & CGF-ACV (Ours) & \textbf{1.03} & \textbf{1.34} & \textbf{0.78} & \textbf{1.01} & 0.4 & 0.5  &\textbf{1.32} &3.08 &\textbf{1.61} &240\\
    % & LEAStereo~\cite{cheng2020hierarchical} &1.90 & \textbf{2.39} & \textbf{1.13} & \textbf{1.45} & 0.5 & 0.5  &\textbf{1.40} &\textbf{2.91} &\textbf{1.65} &300 \\
    \midrule
    \multirow{8}{*}{ \rotatebox{90}{\textit{Speed}}}
    % & {DispNetC\cite{dispNetC} } & 4.11 & 4.65 & 2.77 & 3.20 & 0.9 & 1.0 & 4.32 & 4.41 & 4.34 & 60\\
    % & {StereoNet \cite{stereonet2018}} & - & - & - & - & 0.8 & 0.9 & 4.30 & 7.45 & 4.83 & \textbf{15}\\
     & {DeepPrunerFast\cite{deeppruner}} & - & - & - & - & - & - & 2.32 & 3.91 & 2.59 & 62\\
     & {AANet\cite{aanet}} & 1.91 & 2.42  & 1.46 &1.87 & 0.5 & 0.6 &  1.99 & 5.39 & 2.55 & 62\\
    & DecNet\cite{decomposition}  & - & - & - & - & - & - & 2.07 & 3.87 & 2.37 & 50\\
    % &  {BGNet\cite{bgnet}} & 1.77 & 2.15 &  - & - & 0.6 & 0.6 & 2.07 & 4.74  & 2.51 & 25\\
    &  {BGNet+\cite{bgnet}} & 1.62 & 2.03 &  1.16 & 1.48 & 0.5 & 0.6 & 1.81 & 4.09  & 2.19 & 32\\
    &  {CoEx\cite{coex}} & 1.55 & 1.93 &  1.15 & 1.42 & 0.5 & 0.5 & 1.79 & 3.82  & 2.13 & 27\\
    &  {Fast-ACVNet+\cite{fast-acv}} & 1.45 & 1.85 &  1.06 & 1.36 & 0.5 & 0.5 & 1.70 & 3.53  & 2.01 & 45\\
    &  {HITNet\cite{hitnet}}  & \textbf{1.41} & 1.89 &  1.14 & {1.53} & 0.4 & 0.5 & 1.74 & \textbf{3.20}  & 1.98 & 20\\
    % & CGF-Stereo (Ours) &{1.47} &{1.82} & {1.09} & {1.34} & 0.5 &0.5 & {1.72} & 3.62 & {2.04} & 28 \\
    & CGI-Stereo (Ours) &\textbf{1.41} &\textbf{1.76} & \textbf{1.05} & \textbf{1.30} & 0.5 &0.5 & \textbf{1.66} & 3.38 & \textbf{1.94} & 29 \\
    \bottomrule
    \end{tabular}
    \vspace{10pt}
    \caption{Comparison with the state-of-the-art methods on KITTI benchmarks. \textbf{Bold}: Best.}
    \label{tab:evaluation_kitti}
    \vspace{-10pt}
\end{table*}

\textbf{KITTI 2012 and 2015}. As shown in Tab.~\ref{tab:evaluation_kitti}, our CGI-Stereo ranks $1^{st}$ on KITTI 2012 and 2015 benchmarks among all the published real-time methods. Compared to some real-time methods such as DeepPrunerFast\cite{deeppruner}, AANet\cite{aanet}, and DecNet\cite{decomposition}, our CGI-Stereo not only consistently outperforms them by a considerable margin, but is also faster. More importantly, our method also achieves better performance than HITNet\cite{hitnet}. Our CGF-ACV, which embeds the CGF into ACVNet\cite{acvnet}, ranks $1^{st}$ on the KITTI 2012 and 2015 leaderboards among all the published methods. Qualitative comparisons are shown in Fig.~\ref{fig:cgi_acv}.

\subsection{Generalization Performance}
We compare our methods with several other stereo methods, including the non-real-time methods and real-time methods. In this evaluation, all the comparison methods are only trained on the synthetic Scene Flow\cite{dispNetC} training set, and then evaluated on four real-world datasets, i.e. KITTI 2012\cite{kitti2012} and 2015\cite{kitti2015}, Middlebury 2014\cite{middlebury}, and ETH3D\cite{eth3d}. Tab.~\ref{tab:generalize} summarizes the comparisons. Among all real-time methods, our CGI-Stereo achieves superior generalization performance to others. Furthermore, compared to domain generalized method DSMNet\cite{dsmnet}, our method not only has better generalization performance, but also is $55\times$ faster than it. Qualitative results are shown in Fig.~\ref{fig:genera_kitti} and \ref{fig:genera_midd_eth}.

\subsection{Limitation and Discussion}
In generalization evaluation, the CGF may interfere with disparity prediction for smooth objects when there is rich texture in the context. For example, in the third row of Fig. \ref{fig:genera_midd_eth}, the stop marker on the sign is well visible in the disparity map while the real disparity is quite smooth. For fine-tuning evaluation, the context is better learned, which hardly interferes with the disparity prediction on smooth objects. For example, in the first row of Fig. \ref{fig:cgi_acv}, the `30' marker on the sign is invisible in the disparity and the predicted disparity is quite smooth. We plan to address this limitation in generalization by exploring two potential solutions in the future. One solution is to smooth the context at the object level by Transformer, and then fuse the smoothed context into geometry. Another solution is to use semantic labels to supervise the context to obtain semantically smooth context.

\begin{table} \small
    \centering
    \begin{tabular}{l|cccc}
     \toprule
     \multirow{2}{*}{Method} & \multicolumn{2}{c}{KITTI} & \multirow{2}{*}{Middlebury} & \multirow{2}{*}{ETH3D}\\
     & 2012 & 2015 & & \\
    %  \midrule
    %  CostFilter &21.7 &18.9 & 40.5 &31.1 \\
    %  PatchMatch &20.1 &17.2 &38.6 &24.1\\
    %  SGM &7.1 &7.6 &25.2 &12.9 \\
     \midrule
     PSMNet\cite{psmnet} & 6.0 & 6.3 & 15.8 &9.8 \\
     GANet\cite{ganet} &10.1 &11.7 &20.3 &14.1 \\
     DSMNet\cite{dsmnet} &6.2 &6.5 & 13.8 &6.2\\
     CFNet\cite{cfnet} &5.1 &6.0 &15.4 &5.3 \\
     STTR\cite{sttr} &8.7 &6.7 &15.5 &17.2 \\
     RAFT-Stereo\cite{raft-stereo} &- & 5.7 & 12.6 & \textbf{3.3} \\
     FC-PSMNet\cite{fc} & 5.3 &5.8 &15.1 &9.3 \\
     Graft-PSMNet\cite{graftnet} & \textbf{4.3} &\textbf{4.8} &\textbf{9.7} &7.7 \\
     \midrule
     DeepPrunerFast\cite{deeppruner} &7.6 &7.6 &38.7 &36.8 \\
     BGNet\cite{bgnet} &12.5 &11.7 &24.7 &22.6 \\
     CoEx\cite{coex} &7.6 &7.2 &14.5 &9.0 \\
     % CGF-Stereo (Ours) &6.1 &6.0 &\textbf{11.6} &{6.6} \\
     CGI-Stereo (Ours) &\textbf{6.0} &\textbf{5.8} &\textbf{13.5} &\textbf{6.3} \\
     
    \bottomrule
    \end{tabular}
    \vspace{10pt}
    \caption{Generalization performance on KITTI, Middlebury and ETH3D. All models are only trained on Scene Flow. We split methods into two categories: non-real-time and real-time (from top to bottom).}
\label{tab:generalize}
\vspace{-10pt}
\end{table}

\section{Conclusion}
We have proposed CGI-Stereo, a novel neural network architecture that can concurrently achieves real-time performance, competitive accuracy, and strong generalization ability. We propose CGF to adaptively fuse context into geometry information for more accurate and efficient cost aggregation and meanwhile more effective contextual feature extraction. The proposed CGF can be easily embedded into many existing stereo matching networks, such as PSMNet, GwcNet and ACVNet. The resulting networks are improved in accuracy by a large margin. We hope the proposed CGF can stimulate future research on learning context and geometry for accurate stereo matching.

{\small
\bibliographystyle{ieee_fullname}
\bibliography{egbib}
}

\end{document}